\documentclass[11pt,twocolumn]{article}
\usepackage[a4paper, margin=2.2cm]{geometry}
\usepackage{authblk}
\usepackage{hyperref}
\usepackage[american]{babel}

\usepackage{natbib} 
\usepackage{mathtools} 
\usepackage{amssymb}
\usepackage{bm}
\usepackage{booktabs} 
\usepackage{tikz} 

\newcommand{\R}{\mathbb{R}}
\newcommand{\N}{\mathcal{N}}

\title{Encoding Domain Information with Sparse Priors for Inferring Explainable Latent Variables}

\author[1,2]{Arber Qoku}
\author[1,2,3]{Florian Buettner}
\affil[1]{German Cancer Consortium (DKTK) and German Cancer Research Center (DKFZ)}
\affil[2]{Goethe University Frankfurt, Germany} 
\affil[3]{Frankfurt Cancer Insititute, Germany \authorcr  {\tt \{arber.qoku, florian.buettner\}@dkfz.de}\vspace{1.5ex}} 

\date{}
\begin{document}
\maketitle

\begin{abstract}
Latent variable models are powerful statistical tools that can uncover relevant variation between patients or cells, by inferring unobserved hidden states from observable high-dimensional data. A major shortcoming of current methods, however, is their inability to learn sparse and interpretable hidden states. Additionally, in settings where partial knowledge on the latent structure of the data is readily available, a statistically sound integration of prior information into current methods is challenging.
To address these issues, we propose spex-LVM, a factorial latent variable model with \textbf{s}parse \textbf{p}riors to encourage the inference of \textbf{ex}plainable factors driven by domain-relevant information.
spex-LVM utilizes existing knowledge of curated biomedical pathways to automatically assign annotated attributes to latent factors, yielding interpretable results tailored to the corresponding domain of interest.
Evaluations on simulated and real single-cell RNA-seq datasets demonstrate that our model robustly identifies relevant structure in an inherently explainable manner, distinguishes technical noise from sources of biomedical variation, and provides dataset-specific adaptations of existing pathway annotations.
Implementation is available at \href{https://github.com/MLO-lab/spexlvm}{https://github.com/MLO-lab/spexlvm}.
\end{abstract}

\section{Introduction}
Recent advances in omics profiling technologies from proteomics to single-cell RNA-sequencing (scRNA-seq) provide an ever-finer molecular characterization of patients. In particular, scRNA-seq technologies facilitate the investigation of biological processes at a cellular level. Such a significant increase in data resolution paints a much more comprehensive picture of the underlying drivers of heterogeneity not only between patients, but also at the inter-cellular level. This allows the discovery of new patient sub-populations as well as novel cell sub-populations present across patients, which in turn has the potential to accelerate progress in personalized medicine. However, extracting biologically and clinically meaningful structures from thousands or millions of cells and hundreds of patients presents an array of computational challenges\citep{lahnemann2020eleven}. 
Latent variable modeling is a general approach suitable for finding major axes of variation between samples in terms of unobserved numerical components. However, understanding the output of such models usually involves an ongoing exchange between data analysts and domain experts, which can be inflexible and time-consuming. A much more efficient alternative is taking advantage of established domain knowledge to automatically infer interpretable latent factors, thereby facilitating a faster transition to conclusive insights. Biomedical gene set annotations, for instance, serve as a reliable source of information for encoding a relevant structure of the latent factors, providing an additional layer of understanding to numerically encoded outputs.\\
\begin{figure*}[t]
\centering
\includegraphics[width=\textwidth]{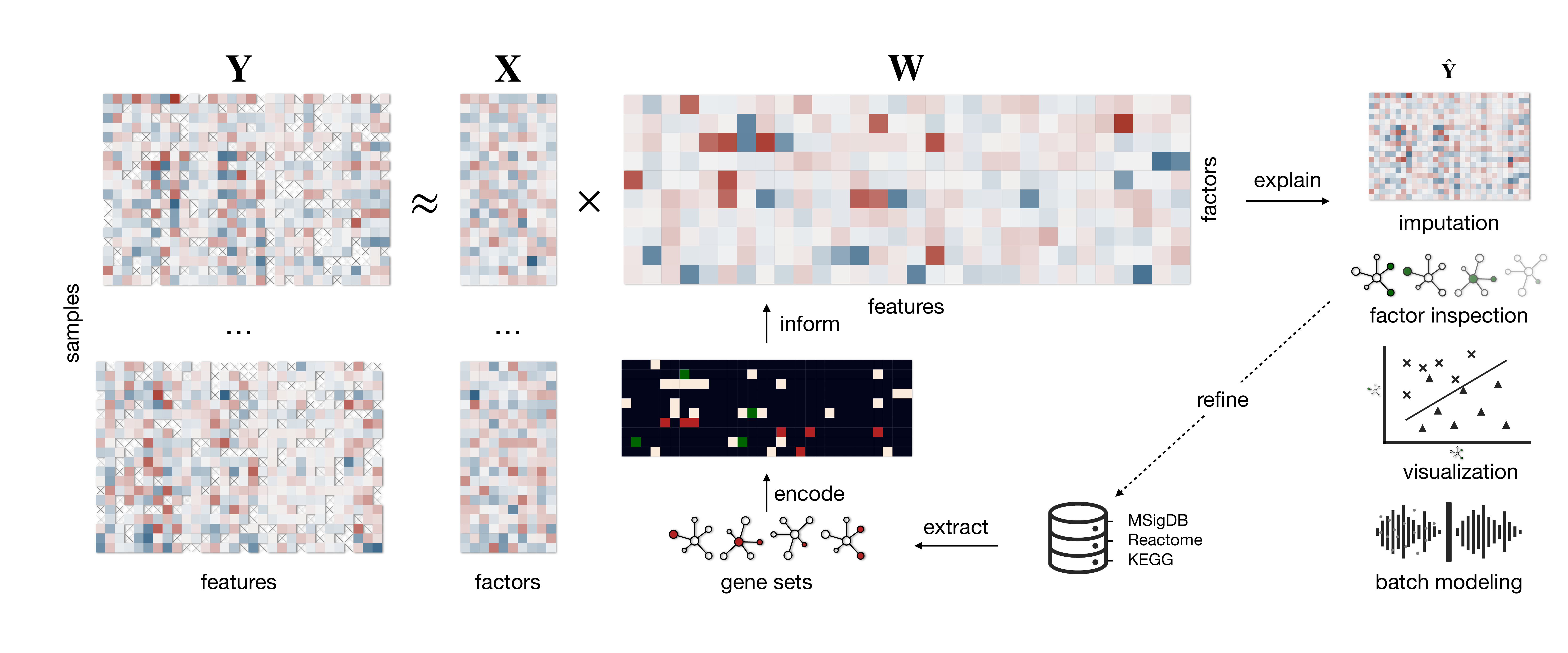}
\caption{Schematic overview of spex-LVM. A gene expression matrix is decomposed as a sum of latent factors informed by potentially noisy gene set annotations. Spex-LVM learns the true underlying axes of variation by pruning redundant factors and refining noisy annotations. On the right we depict a typical downstream analysis involving imputation of incomplete observations, factor ranking and inspection of refined features, mapping of samples onto latent axes for visualization and modeling of technical noise via additional unannotated factors.}
\label{fig:teaser}
\end{figure*}
We propose spex-LVM, a flexible and efficient latent variable model capable of utilizing prior information of well-known biomedical gene set annotations to guide the inference of explainable sources of variation in scRNA-seq data.
More specifically, spex-LVM learns interpretable factors that capture meaningful subpopulations of cells or patients, automatically tagged with the biomedical pathway characterizing the corresponding axis of variation. We account for unwanted technical noise\citep{hicks2018missing} by explicitly modeling irrelevant sources of variation as additional factors that capture perturbations affecting large groups of features. In addition, our approach considers potentially noisy annotations and provides refined alternatives supported by the data. 
Finally, our method provides identifiable latent factors. Experiments show that the order of the learned factors matches the order of the provided annotations, thereby circumventing a known drawback inherent to the conventional factor analysis\citep{shapiro1985identifiability}, and yielding consistent and reproducible results. We should note that our method is not limited to biomedical data analysis, and it can be easily applied to other domains where partial information on the latent structure is available.
\section{Method}
Spex-LVM decomposes high dimensional gene expression data into distinct factors that jointly contribute to the data generation process. One major advantage of our model is its ability to incorporate existing knowledge of biomedical pathways to encourage learning of interpretable factors as depicted schematically in Figure~\ref{fig:teaser}. Furthermore, the model assumes incomplete prior knowledge by modeling additional unannotated sparse and dense factors, which serve as placeholders for encoding sparse unidentified pathways, and external or technical perturbations that affect the expression pattern of nearly all genes.\\
Let $\mathbf{Y}\in\R^{N\times G}$ denote the gene expression matrix comprising $N$ samples across $G$ genes. spex-LVM decomposes $\mathbf{Y}$ into a sum of $A$ annotated and $H$ unannotated factors, where the latter can be further differentiated between sparse and dense unannotated factors. 
\begin{equation}
\label{eq:fa_sum}
    \mathbf{Y} = \sum_{a}^{A}\mathbf{p}_a \otimes \mathbf{v}_a + \sum_{h}^{H}\mathbf{q}_h \otimes \mathbf{u}_h + \mathbf{\Psi},
\end{equation}
where $\mathbf{p}_a$, $\mathbf{q}_h \in \R^N$ describe the state of the annotated factor $a$ and the unannotated factor $h$ across $N$ cells, and $\mathbf{v}_a$, $\mathbf{u}_h \in \R^G$ are the corresponding regulatory weights or factor loadings for factors $a$ and $h$ across $G$ genes in $\mathbf{Y}$. The residuals are denoted by $\mathbf{\Psi} = diag(\bm{\sigma}^2)$, a diagonal matrix storing the marginal variances $\sigma_g^2$ of each gene $g$. We may rewrite \ref{eq:fa_sum} concisely in matrix notation by defining $\mathbf{X} \in \R^{N \times K}$ as $\left[\mathbf{p}_1, \cdots, \mathbf{p}_A, \mathbf{q}_1, \cdots \mathbf{q}_H\right]$ and $\mathbf{W} \in \R^{G \times K}$ as $\left[\mathbf{v}_1, \cdots, \mathbf{v}_A, \mathbf{u}_1, \cdots \mathbf{u}_H\right]$.
\begin{equation}
\label{eq:fa_mat}
    \mathbf{Y} = \mathbf{X} \cdot \mathbf{W}^T + \mathbf{\Psi}
\end{equation}
Factor analysis models\citep{bishop2006pattern, murphy2012machine} typically assume the projection of a sample $x_{i, k}\sim\N(0,1)$ in the latent space, to a sample $y_{i, g} \sim\N(\sum_{k}w_{g, k}x_{i, k}, \sigma_g^2)$ in the original high-dimensional space via the corresponding factor loadings $ w_{g, k} \sim \N(0, 1)$.
We extend the conventional factor analysis by introducing a \emph{global-local} shrinkage prior\citep{polson2010shrink} on the scales of each weight $w_{g, k}$. 
\begin{equation}
    w_{g, k} | \tau, \lambda_{g, k} \sim \N(0, \tau^2 \cdot \lambda_{g, k}^2)
\end{equation}
The consequences of this addition are twofold. First, the global scale $\tau$ shrinks all the weights towards zero, encouraging the model to learn sparse representations of the underlying factors. This is especially useful when learning sparse unannotated factors. Second, the local scales act as a regulator on each individual weight. A small local scale $\lambda_{g, k}$ confines $w_{g, k}$ to values very close to zero. On the other hand, a large local scale allows its corresponding parameter to escape the global regularization penalty. As an enhancement to the original horseshoe prior\citep{carvalho2009handling}, the regularized version provides means to control the slab width of each individual scale\citep{piironen2017sparsity}, allowing partial knowledge to guide the model, e.g.\ annotated gene sets.
\begin{align}
w_{g, k} | \tau, \lambda_{g, k}, c_{g, k} &\sim \N(0, \tau^2 \cdot \tilde{\lambda}_{g, k}^2),\\
\tilde{\lambda}_{g, k} &= \frac{c_{g, k}^2\lambda_{g, k}^2}{c_{g, k}^2 + \tau^2\lambda_{g, k}^2}
\end{align}
The slab width effectively encodes our confidence about the given gene set annotations. 
We accommodate potentially noisy annotations by providing soft bounds for the local scales, allowing the adaptation of incorrect signals given sufficient evidence from the data. 
Experiments show that a slab width of $c_{g, k} \approx 1.0$ for genes annotated to a pathway and $0 < c_{g, k} < 0.1$ for unannotated genes provide the best results.
We assign a relatively large slab width across all genes for the unannotated dense factors, and vice-versa for the unannotated sparse factors, allowing the model to autoregulate active signals based on the data.
Alternative implementations of sparse factor analysis models rely on the spike-and-slab prior \citep{buettner2017f,mitchell1988bayesian}; however, due to their discrete nature, these models cannot be trained using blackbox stochastic variational inference and therefore tend to be inefficient. 
We implement the alternative parameterization of the regularized horseshoe as suggested in \citep{piironen2017sparsity} and described above. The resulting algorithm is amenable to efficient and scalable stochastic variational inference\citep{ranganath2014black, hoffman2013stochastic, wingate2013automated}, and optimizes the evidence lower bound via standard gradient methods such as Adam\citep{kingma2014adam}.
\section{Results and Discussion}
We first evaluate several properties of spex-LVM on synthetic data. We show that spex-LVM learns informed and relevant factors, while redundant annotated factors that do not explain variation in a specific dataset are pruned. Second, we illustrate that spex-LVM is robust against noisy gene set annotations. Third, we demonstrate that spex-LVM scales sub-linearly with the number of samples and handles millions of samples and tens of thousands of genes.\\
We then show on three real-world datasets that spex-LVM explicitly models and separates unwanted variation from identified pathway factors that are biologically meaningful. 
For each experiment, we derive gene set annotations from established public databases such as MSigDB\citep{liberzon2015molecular} and Reactome\citep{fabregat2018reactome}.
\subsection{Synthetic Data}
\begin{figure}[h]
\centering
\includegraphics[width=\linewidth]{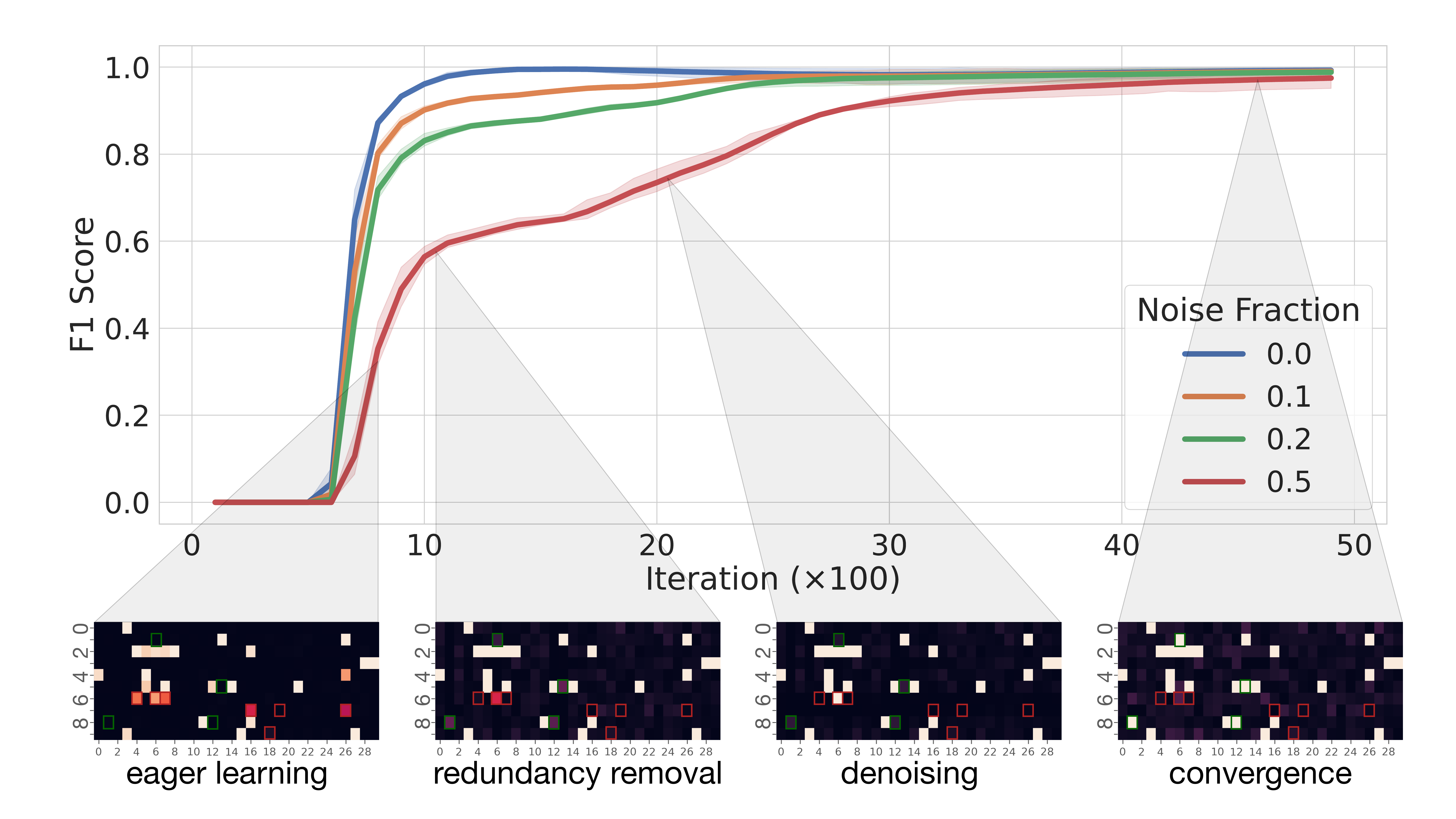}
\caption{Learning from noisy information. Models are trained independently on different levels of deliberate noise. Each line summarizes the progress of the $F_1$ score computed from the true annotation signals before introducing noise and the learned signals. Below, a subset of the learned weights after specific iterations as heatmaps. Colors encode absolute gene weight (black: inactive, white: active). Noisy factor loadings are indicated with colored edges (red: false positives, green: false negatives).}
\label{fig:noise}
\end{figure}
\begin{figure}[h]
\centering
\includegraphics[width=\linewidth]{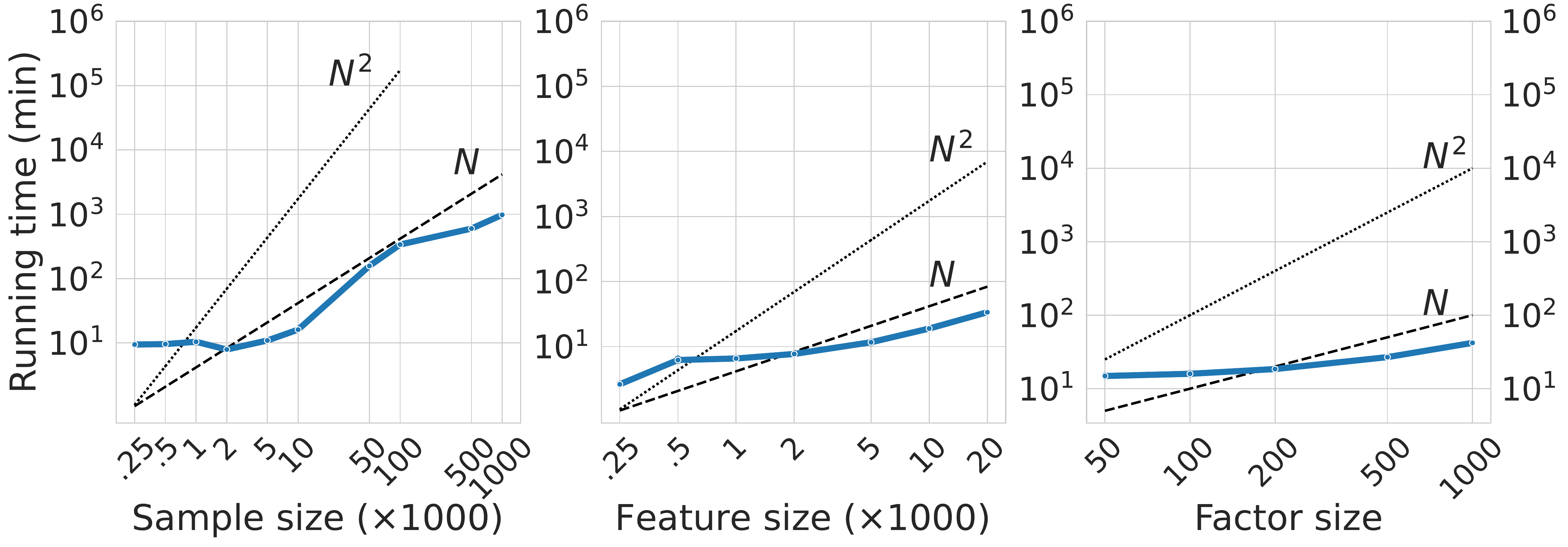}
\caption{Scalability of spex-LVM in terms of sample, feature and factor size. The $y$-axes match on all plots to help gauge the relative impact of each data dimension.}
\label{fig:runtimes}
\end{figure}

\noindent
We assess the ability of our model to discern valuable prior information from noisy annotations. We simulate a gene expression toy dataset of $N=$ 10,000 cells and $G=$ 8,000 genes as a linear transformation of $K=100$ latent factors generated from a standard normal distribution $\N(0, \mathbf{I})$, where $\mathbf{I}$ denotes the $K$-dimensional identity matrix. The factor loadings were generated from a zero-centered normal distribution $\N(0, 4)$, where loadings with an absolute value of less than $0.5$ were set to zero, to emphasize the gap between active and inactive signals. In addition, we randomly set $85\%$-$95\%$ of the loadings to zero to more closely resemble the true size of gene set collections. After extracting the true factor mask, indicating active and inactive genes for each factor, we introduce additional noise in the form of false positive, false negative annotations and redundant factors. Figure~\ref{fig:noise} shows a summary of the results for several models trained independently on different noise fractions. A noise fraction of $0.1$ means that $10\%$ of true active genes have been switched off, i.e.\ false negatives, as well as $10\%$ of true inactive genes have been incorrectly switched on, i.e.\ false positives. For each noise level we train four different models with $0\%$, $10\%$, $20\%$ and $50\%$ additional random factors, irrelevant to the data generation process. The validation on the toy data demonstrates that spex-LVM learns relevant structure from noisy information and converges to the true latent factors. We identify three distinct stages during model training (Fig.~\ref{fig:noise}). In stage one, the model eagerly discovers all of the factors provided by the annotation mask, as indicated by multiple false positive factor loadings illustrated with red colored edges in the first heatmap. The redundant factors 6 and 7 are weakly identified, and quickly vanish during the second learning stage. We call the third stage the denoising stage as the model iteratively flips false positives and false negatives into their correct state. The fourth and last heatmap in Fig.~\ref{fig:noise} shows the state of the weights after the model has fully converged.\\
We conclude the synthetic evaluation by providing a summary of the running times (Fig.~\ref{fig:runtimes}) recorded across different dataset sizes with respect to the number of samples, features and latent factors. While sweeping one parameter, the default values for the other parameters are as follows: $N=$ 10,000 samples, $D=$ 8,000 features and $K=$ 100 factors. Models were trained on a single NVIDIA Quadro RTX 5000 GPU with 16 GB memory.
The available memory can fit up to 20 thousand samples by approximately 10 thousand features concurrently. The significant jump in the elapsed time for the data with more than 20 thousand samples is due to the introduction of batches in the training process. Consequently, the number of samples presents the largest impact in the amount of time required for the algorithm to converge. However, for more than 100 thousand samples, we see a sub-linear increase in the runtime as the number of the total batch updates necessary for convergence does not increase as rapidly with the sample size. On the other hand, due to the parallelized computation, spex-LVM efficiently handles up to 20 thousand features and fit hundreds or even thousands of latent factors for tens of thousands of samples in less than 30 minutes on average.
\subsection{Mouse Embryonic Stem Cells}
\begin{figure}[h]
\centering
 \includegraphics[width=0.85\linewidth]{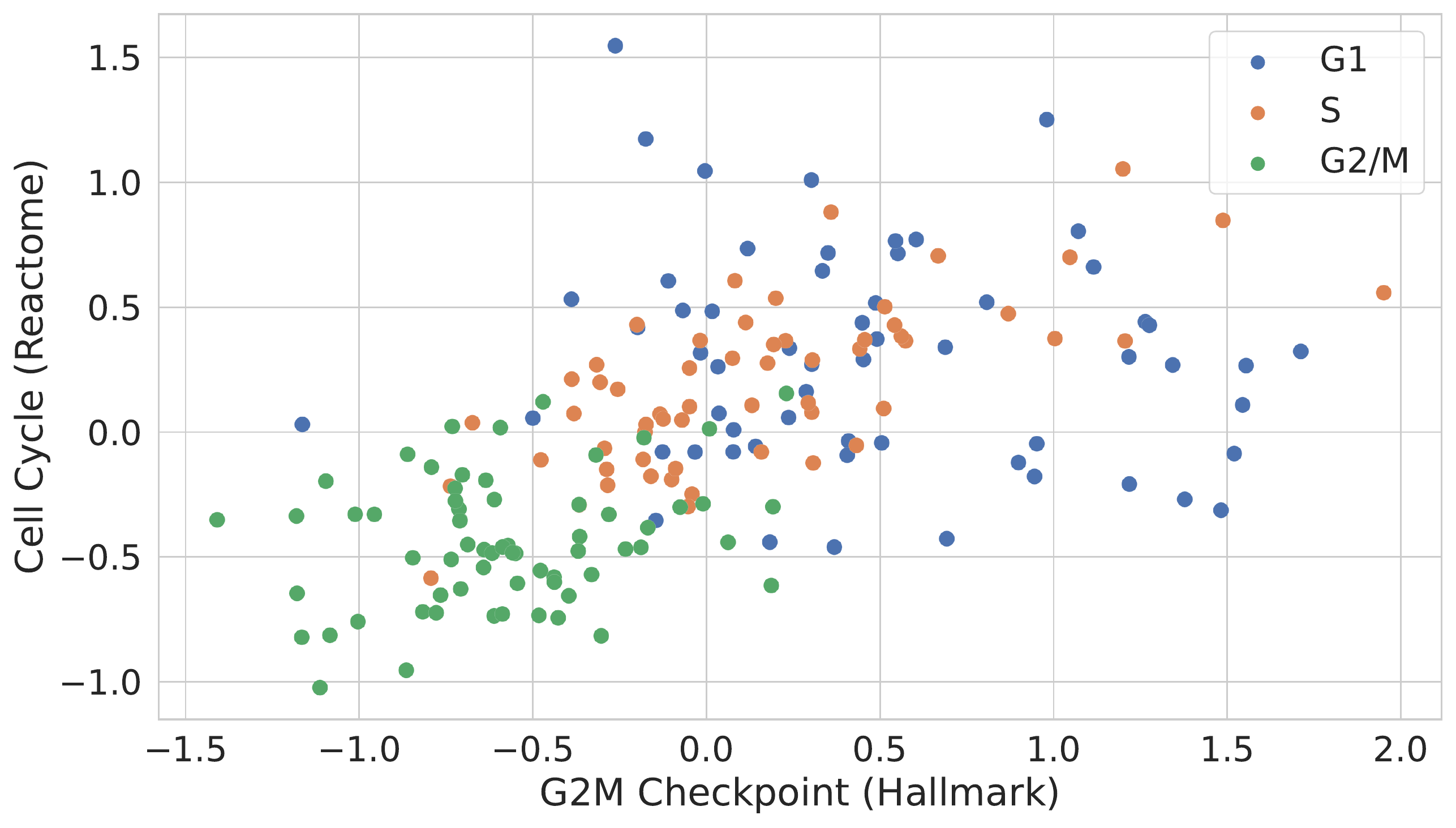}
\caption{Bivariate visualization of mouse embryonic stem cells in terms of the two main factors, the G\textsubscript{2}/M checkpoint (H) and the cell cycle (R). Each cell is color-coded according to its corresponding stage in the cell cycle.}
\label{fig:mesc}
\end{figure}

\noindent
We evaluate our model on 182 mouse embryonic stem cells transcriptomes measured across 6 thousand genes, where the stage of the cell cycle is determined a priori and provided for each cell\citep{buettner2015computational}. During training, we provide annotations from 29 hallmark gene sets and 54 reactome gene sets with a median size of 68 genes. As expected, the model robustly identifies biological pathways responsible for the cell cycle. Cell cycle (R), G\textsubscript{2}/M checkpoint (H) and P\textsubscript{53} pathway (H) comprise three of the top five candidates in terms of variance explained. 
Figure~\ref{fig:mesc} depicts the cell population mapped onto the latent axes corresponding to the G\textsubscript{2}/M checkpoint and the cell cycle.
While both factors contribute in finding sub-populations of cells during the same cell cycle phase, the G\textsubscript{2}/M checkpoint is a stronger indicator of cells in the G\textsubscript{2}/M phase as supported by their corresponding area under the receiver operating characteristic curve (\emph{AUROC}) of 0.76 and 0.82, respectively.

\subsection{PBMCs from Lupus Patients}
\label{subsec:pbmc}
\begin{figure}[h]
\centering
  \begin{minipage}{0.52\linewidth}
    \includegraphics[width=\linewidth]{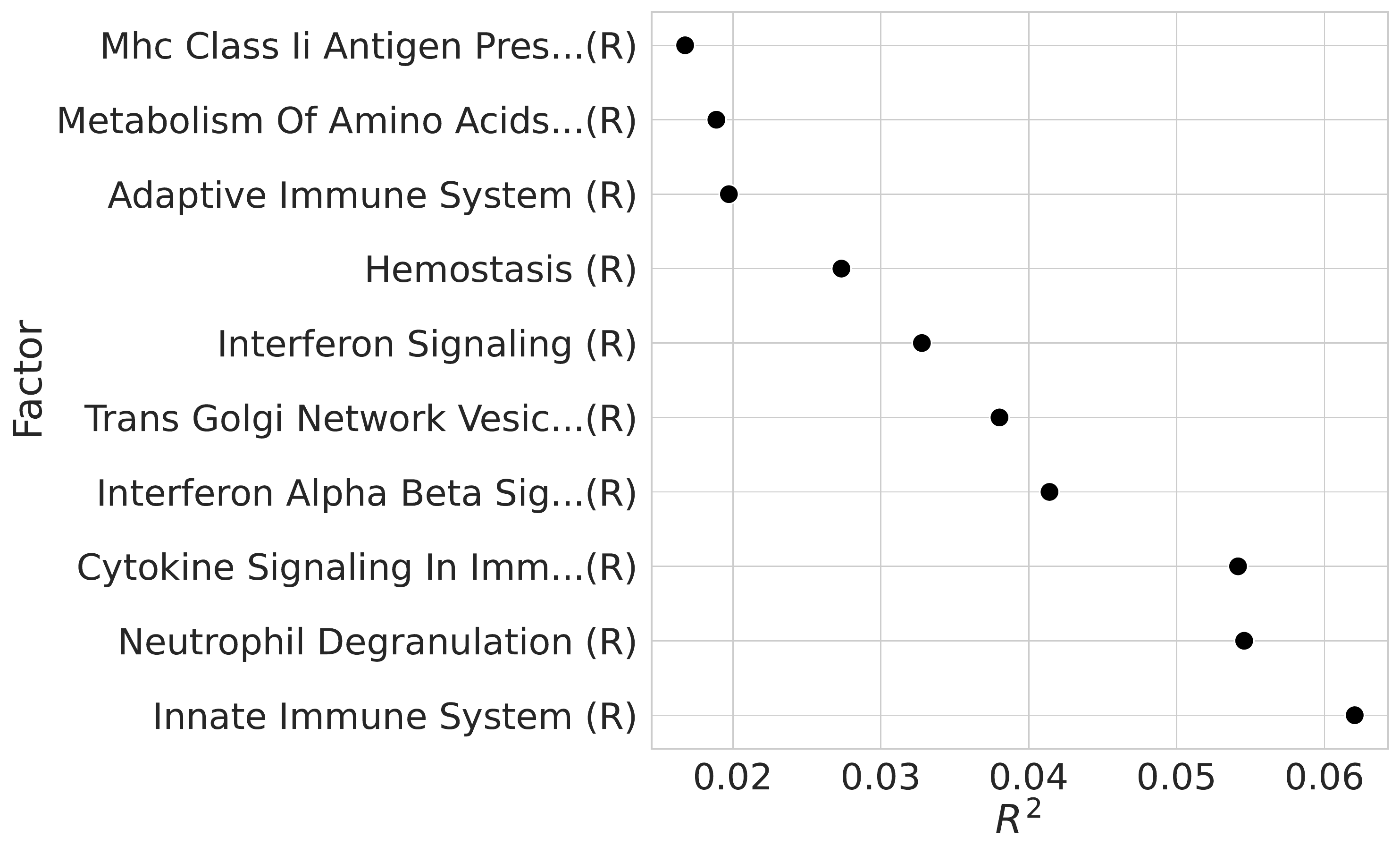}
  \end{minipage}%
   \hfill
  \begin{minipage}{0.44\linewidth}
    \includegraphics[width=\linewidth]{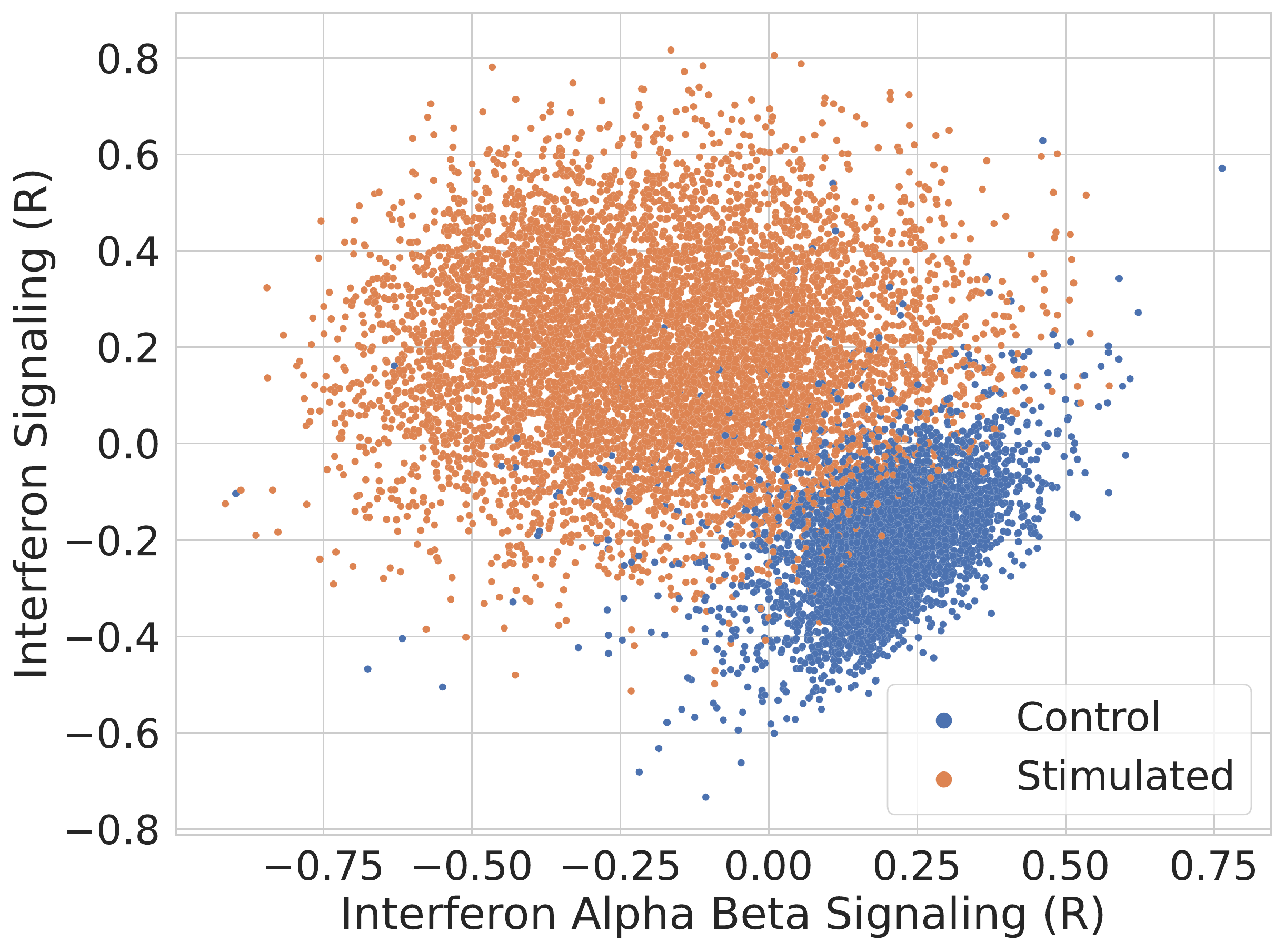}
  \end{minipage}%
\caption{PBMCs from lupus patients results. Top 10 most relevant factors discovered by spex-LVM (left). Two interferon signaling related factors capture the differences between the expression profiles of the control group and IFN-$\beta$ stimulated group (right).}
\label{fig:kang_main}
\end{figure}

\begin{figure}[h]
\centering
 \includegraphics[width=\linewidth]{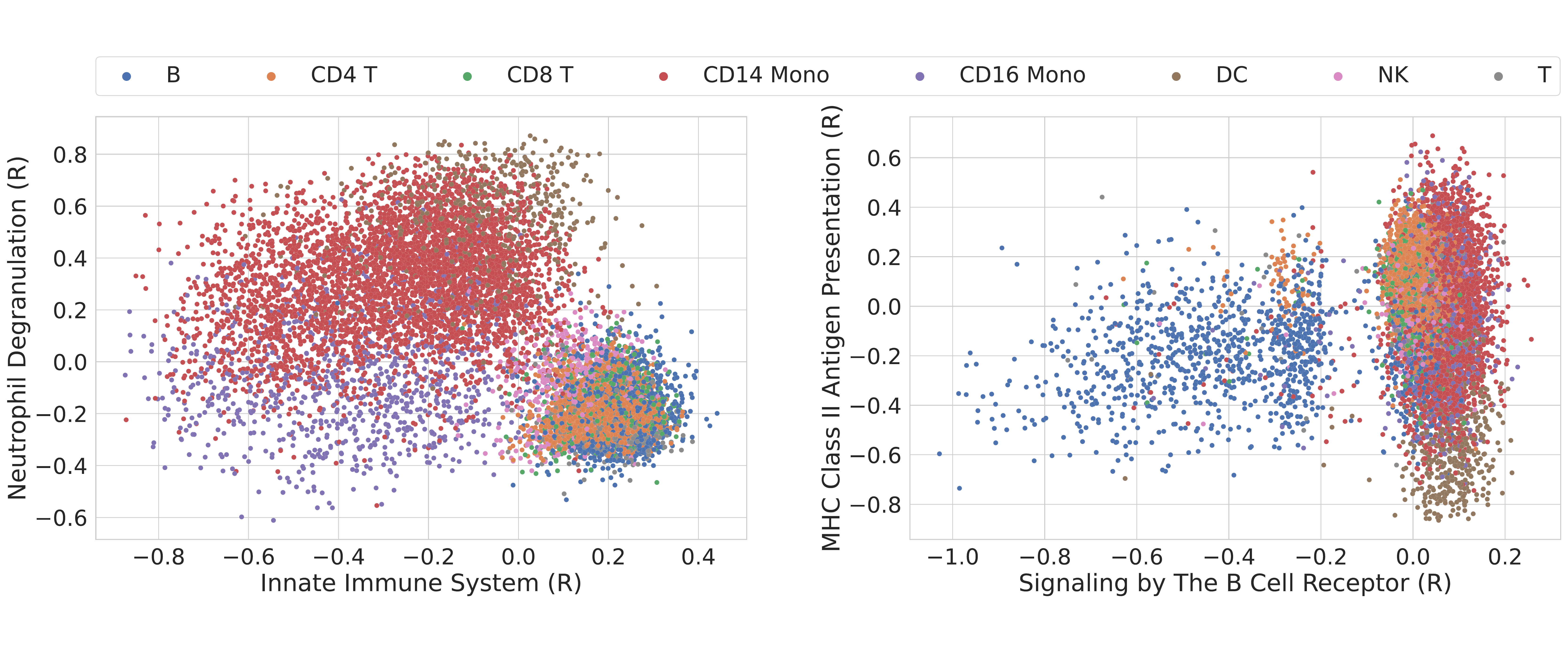}
\caption{Relevant factors related to the immune system response. Factors of general pathways, e.g.\ innate immune system, neutrophil degranulation, explain variation within the larger clusters of monocytes (left). More specific pathways infer axes of variation for smaller clusters such as the dendritic cells and the B cells (right).}
\label{fig:kang_immune}
\end{figure}

Next, we investigate a much larger dataset of peripheral blood mononuclear cells from lupus patients profiled from droplet single-cell RNA-sequencing (dscRNA-seq)\citep{kang2018multiplexed}, where cells were either stimulated with Interferon-$\beta$ or left untreated, yielding two similarly sized groups of cells. We adopt the preprocessing steps from \citep{rybakov2020learning}, and train our model on 13,576 samples, each comprising 979 features or genes. As prior information we consider annotations provided by the reactome gene sets, where at least 15 genes were present in the data, resulting in 131 annotations with a median size of 25. In Figure~\ref{fig:kang_main} we summarize the top 10 most relevant factors based by their corresponding relevance in terms of the coefficient of determination of each factor for the complete dataset.
\noindent
We mainly observe pathways related to immune system and interferon response. In particular, interferon signaling and interferon alpha beta signaling capture the transcriptomic differences between the control and IFN-$\beta$ stimulated group, as anticipated.
Pathways involved in the immune response explain the main drivers of variation along the different cell types as shown in Figure~\ref{fig:kang_immune}.
\noindent
The two larger clusters of monocytes, CD14 and CD16, which comprise nearly $40\%$ of the samples, exhibit significant differences from the rest of the population in terms of the innate immune system and the neutrophil degranulation.
In addition, we observe relatively smaller, yet more specific pathways, which explain less represented cell types. MHC class II antigen presentation describes the process of binding and transporting antigenic peptides to the membrane of professional antigen-presenting cells (APCs), such as dendritic cells\citep{roche2015ins}. Our experiments show that the corresponding factor reliably separates dendritic cells from the other cell types with an \emph{AUROC} score of $0.95$. 
Another candidate is the signaling by the B cell receptor, consisting of only 19 gene set annotations, which consistently infers the axis of variation along the B cells ($0.87$ \emph{AUROC} score).

\subsection{Adult Human Retina}
\begin{figure}[h]
\centering
  \begin{minipage}{0.58\linewidth}
    \includegraphics[width=\linewidth]{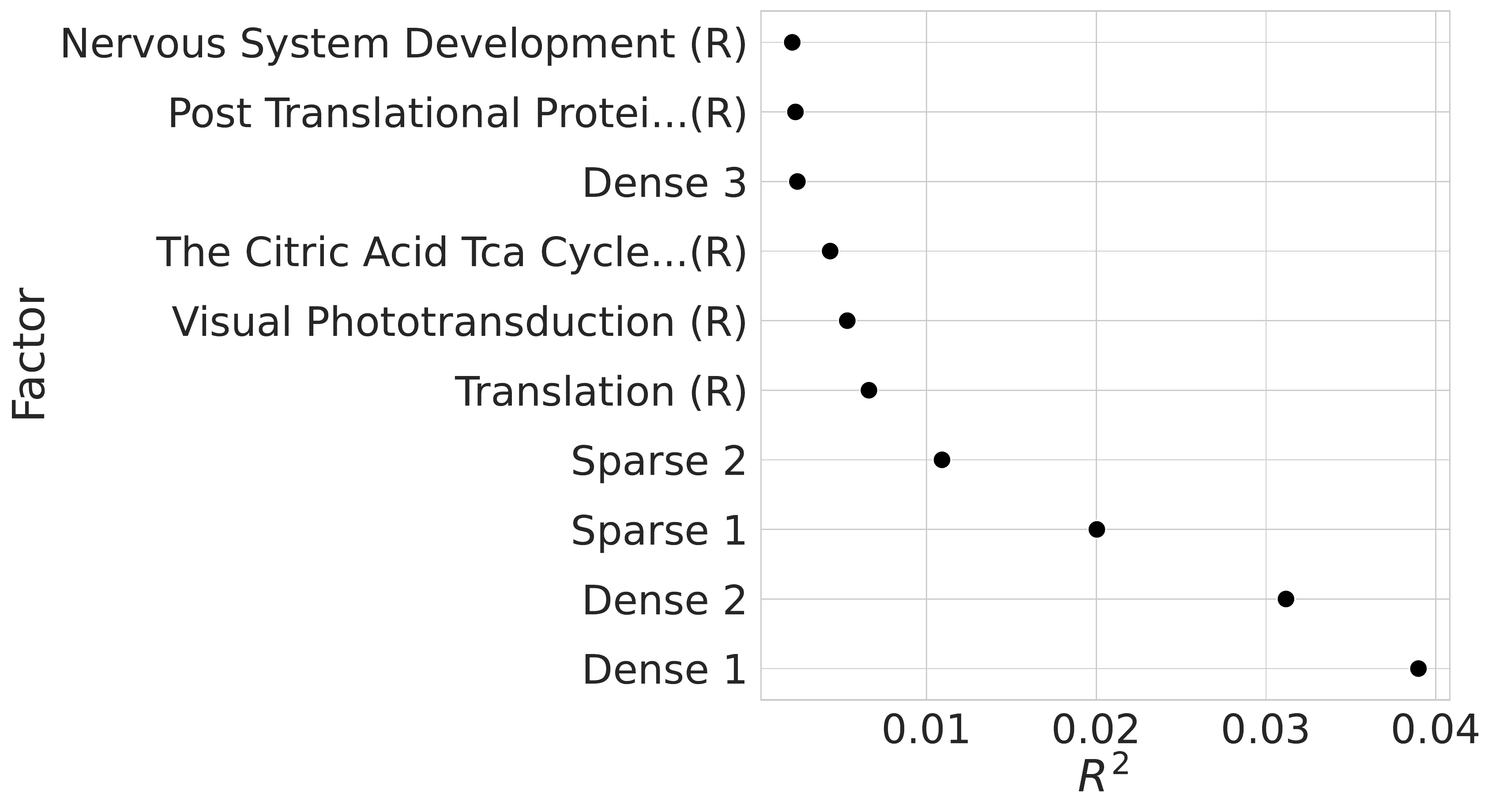}
  \end{minipage}%
   \hfill
  \begin{minipage}{0.37\linewidth}
    \includegraphics[width=\linewidth]{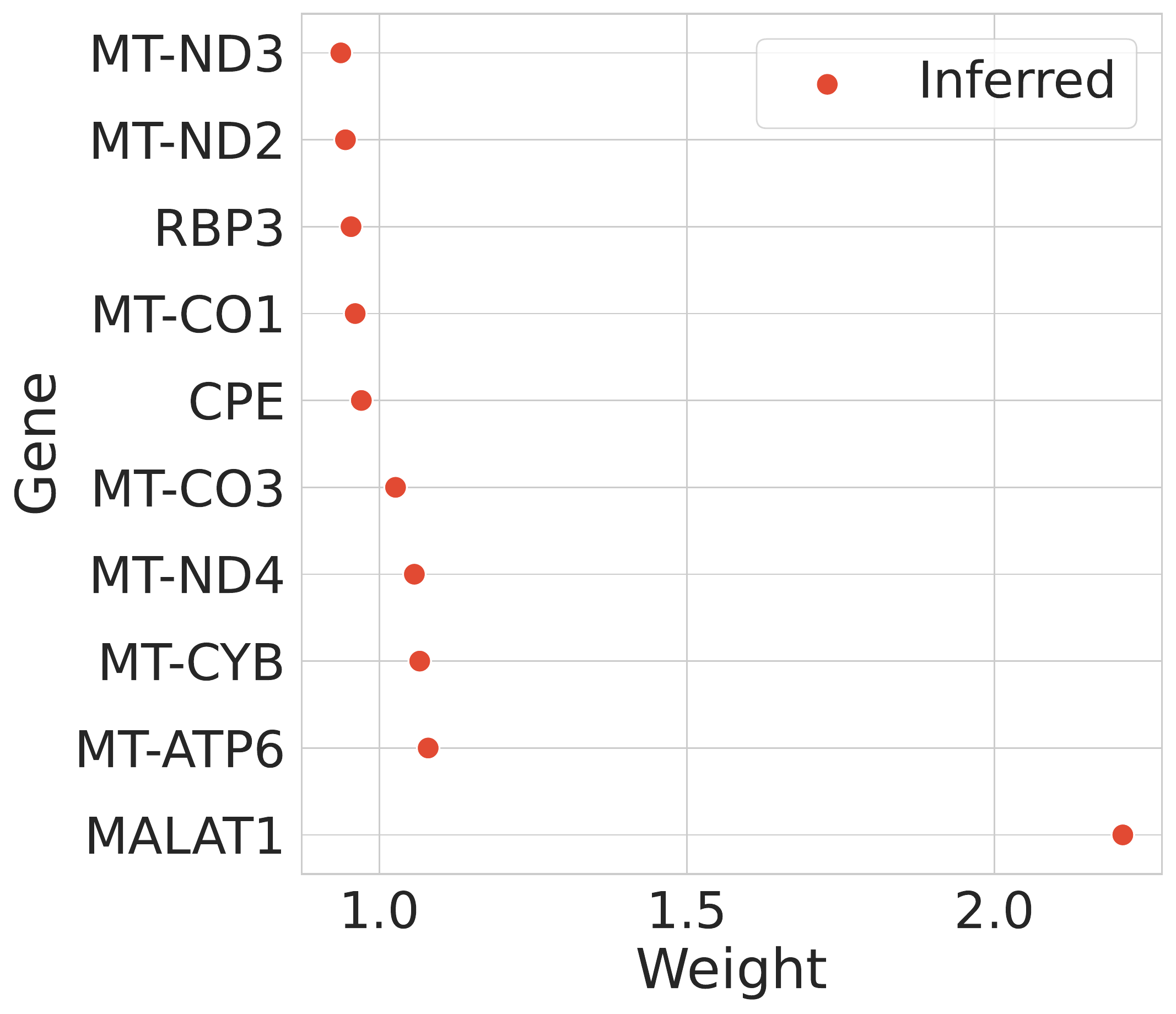}
  \end{minipage}
\caption{Adult human retina atlas results. Top 10 most relevant factors discovered by spex-LVM (left). Sparse factor 1 explains unwanted variation characterized by upregulated low-quality \emph{MALAT1} reads (right).}
\label{fig:retina_large_main}
\end{figure}

Finally, we investigate a transcriptome atlas of the human retina\citep{lukowski2019single}. The dataset consists of 20 thousand cells collected from the neural retina layers of three different donors, and belong to seven transcriptionally distinct clusters. We retain the top 8 thousand most variable features during training.
Due to the increased feature size of the dataset, we derive 35 annotations from hallmark and 161 from reactome gene set collections. Additionally, we allow the model to infer 2 unannotated sparse and 2 dense factors, bringing the total number of latent factors to 200. The results in Figure~\ref{fig:retina_large_main} reveal several insights in accordance with previous findings. 
\noindent
First, the unannotated factors clearly explain the majority of the variance in the data as supported by their accumulative $R^2$ score of approximately $13\%$, which we attribute to potential sources of technical heterogeneity. Sparse factor 1 represents unwanted technical variation in form of an upregulation for low-quality \emph{MALAT1} and mitochondrial reads, as it consists of a highly expressed \emph{MALAT1} gene followed by amplified signals of mitochondrial-encoded genes (\emph{MT-ATP6}, \emph{MT-CYB}, \emph{MT-ND4}, \emph{MT-CO3})\citep{ilicic2016classification}.
\begin{figure}[h]
\centering
  \begin{minipage}{0.38\linewidth}
    \includegraphics[width=\linewidth]{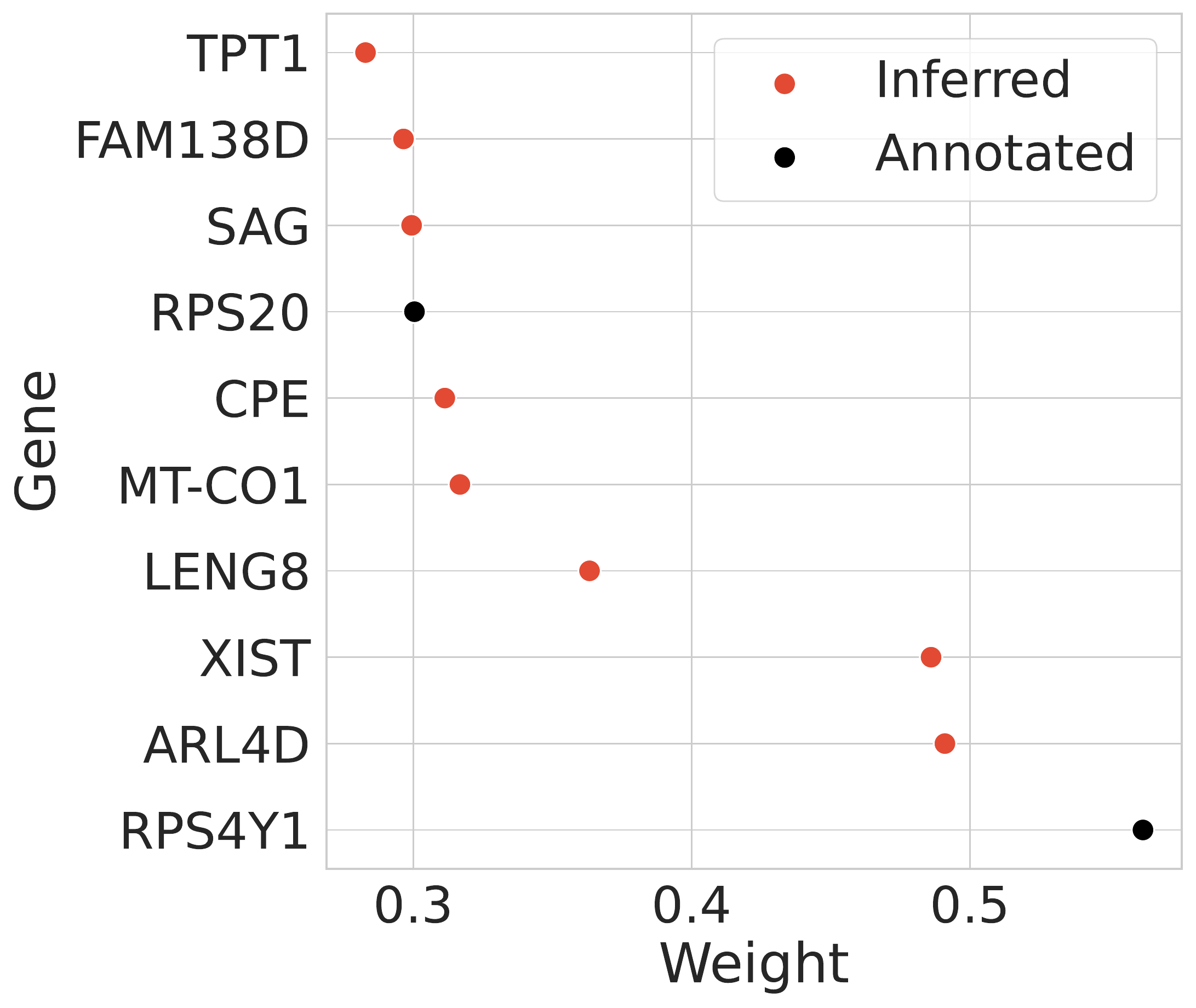}
  \end{minipage}%
   \hfill
  \begin{minipage}{0.6\linewidth}
    \includegraphics[width=\linewidth]{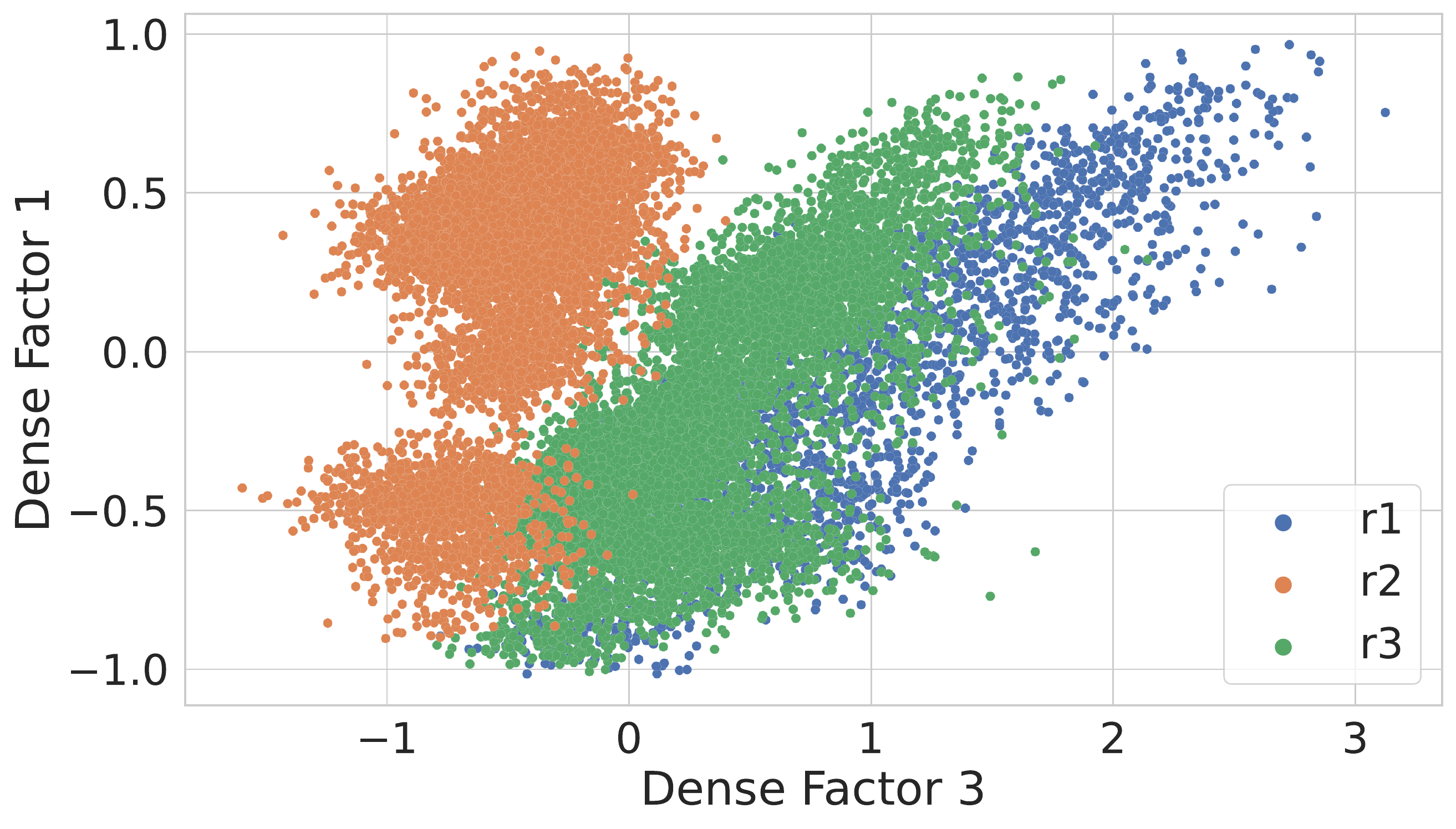}
  \end{minipage}%
\caption{Metabolism of RNA (R) overwritten significantly by the data (left). The new dense factor captures the major axes of variation pertaining to individual donors (right).}
\label{fig:retina_large_batch}
\end{figure}
\noindent
An additional dense factor is introduced during training, as a large portion of the prior annotations of the reactome gene set metabolism of RNA was refined significantly by the model. The new dense factor explains the variation introduced by the different batches, revealing the corresponding donor that provided the data as shown in Figure~\ref{fig:retina_large_batch}.
\begin{figure}[h]
\centering
  \begin{minipage}{0.38\linewidth}
    \includegraphics[width=\linewidth]{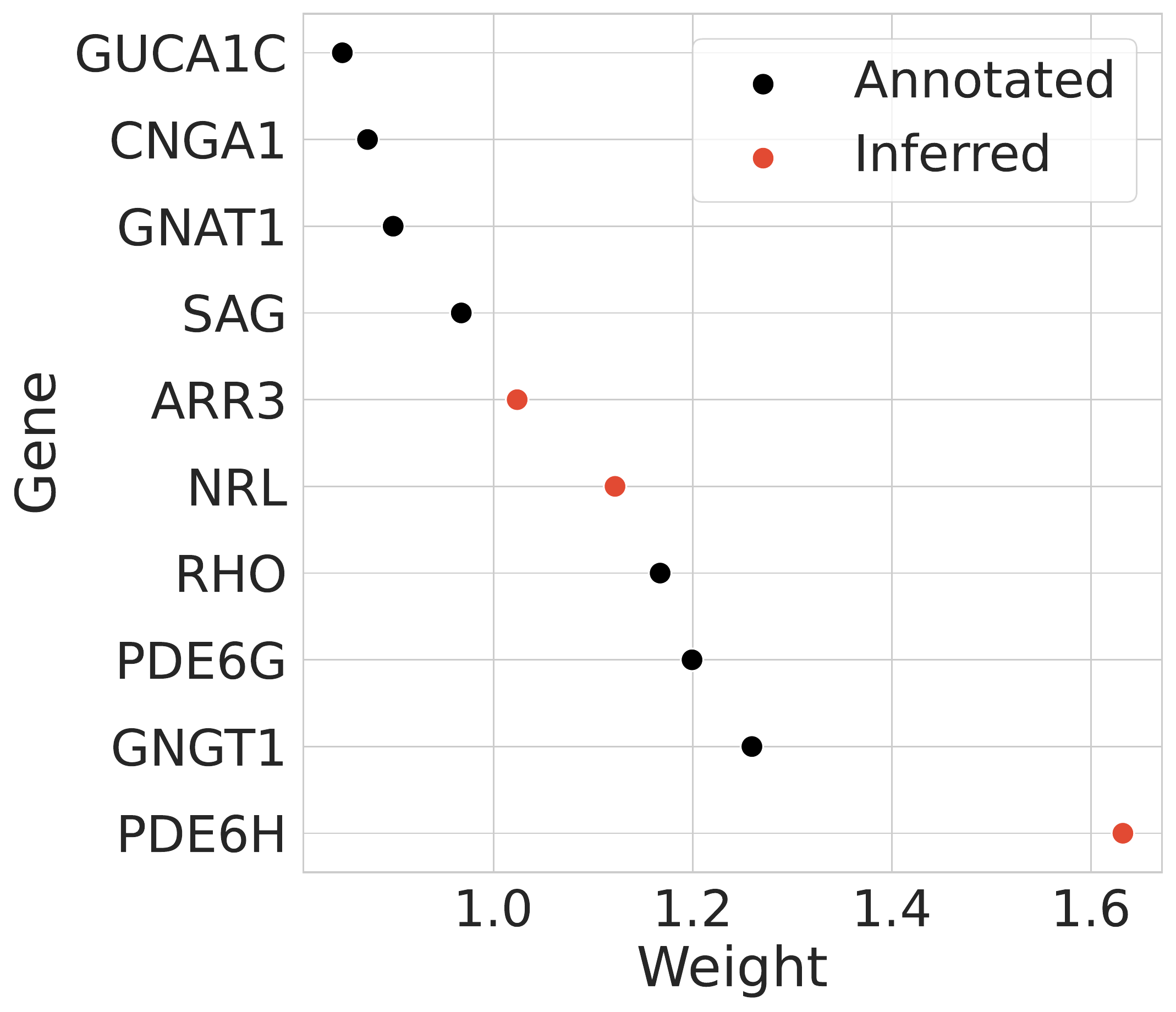}
  \end{minipage}%
   \hfill
  \begin{minipage}{0.59\linewidth}
    \includegraphics[width=\linewidth]{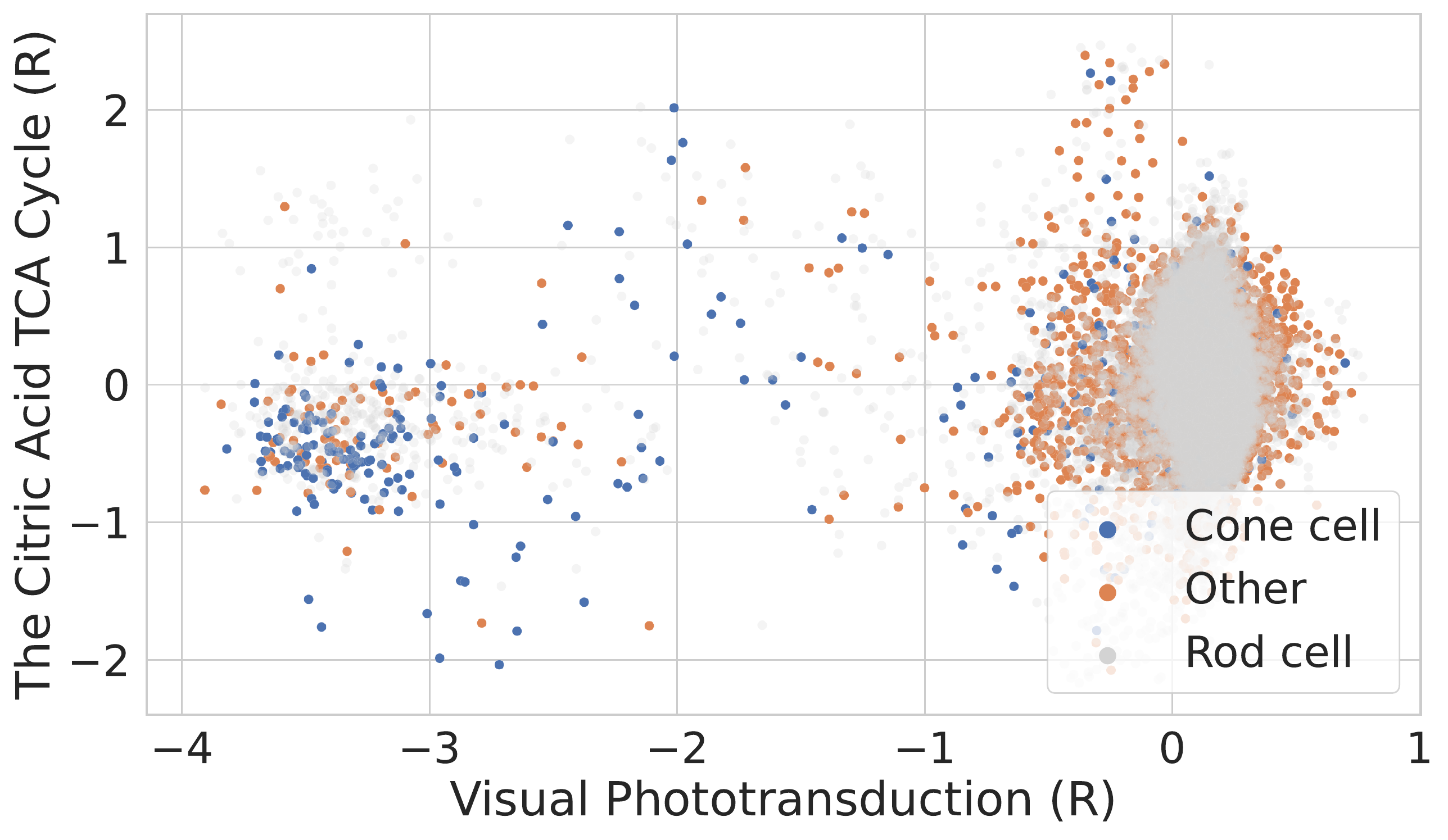}
  \end{minipage}%
\caption{Reactome visual phototransduction factor. Top 10 absolute weights color-coded emphasize added genes (left). Refined visual phototransduction factor serves as a strong indicator of cone photoreceptors (right). Rod cells have altered transparency ($\alpha=0.5$), due to their large number in the dataset.}
\label{fig:retina_phototrans_vs_tca}
\end{figure}
\noindent
Among the annotated factors, we observe active pathways that are responsible for critical processes occurring inside the cell such as the translation of mRNA sequences for protein synthesis, or the citric acid TCA cycle, also known as the Krebs cycle. Importantly, the model consistently identifies factors characterized by specific processes such as visual phototransduction (R) - i.e. the conversion of light into electrical signals in rod and cone photoreceptors - as a highly relevant factor. Although rod photoreceptors dominate the dataset with over 15 thousand $(75\%)$ samples, the original factor is refined by cone specific markers such as \emph{PDE6H} and \emph{ARR3}\citep{lukowski2019single}, rendering the inferred factor also suitable for explaining the variation within the significantly smaller cluster of $627$ cone cells as illustrated in Figure~\ref{fig:retina_phototrans_vs_tca}.
\section{Conclusion}
We proposed spex-LVM, a flexible and interpretable factor analysis approach that supports domain-specific interpretation of discovered latent factors. By leveraging an efficient stochastic variational inference algorithm we demonstrated the scalability of our model to high-dimensional data.
Multiple experimental scenarios on synthetic data showed that our model consistently pinpoints the true axes of variation, regardless whether the prior information was severely perturbed. As an additional benefit, spex-LVM learns identifiable factors as the designated order of the annotated factors is preserved. Further validation of our model on real datasets demonstrated its ability to uncover known sources of heterogeneity, isolate unwanted technical noise, and refine existing gene set annotations in a data driven manner.\\
In the following, we address some limitations of our work and potential future extensions.
First, while linear models promote interpretability, their limited expressiveness cannot capture complex, non-linear dependencies. Unsupervised approaches in deep learning such as the variational autoencoders (VAEs)\citep{kingma2013auto} provide an attractive alternative to traditional linear models, but risk being opaque in favor of more expressive power, and may fail in settings where data samples are scarce.\\
Second, a basic understanding of the domain knowledge is essential to determining the proper amount of bias that should be introduced by the prior information during training. The balance between the prior belief and the likelihood based on the data is a much more complex issue that is prevalent in most statistical methods, and especially in approaches that integrate prior knowledge which may be limited, or otherwise noisy.\\
As future work, the ideas presented in this work can be extended to accommodate multiple views, e.g.\ GFA\citep{klami2014group}, MOFA\citep{argelaguet2018multi}, which would utilize the prior information of a subset of views to indirectly provide explanatory labels for all the modalities involved in the analysis, thereby facilitating the discovery of novel feature sets.




\bibliography{spex-bib}

\end{document}